\documentclass{article} 
\usepackage[preprint]{colm2026_conference}

\usepackage{microtype}
\usepackage{hyperref}
\usepackage{url}
\usepackage{booktabs}
\usepackage{array}
\usepackage{amsmath}
\usepackage{amssymb}
\usepackage{amsthm}
\usepackage{mathtools}
\usepackage{graphicx}
\usepackage{tikz}
\usetikzlibrary{arrows.meta,positioning,calc}
\usepackage{lineno}
\usepackage{enumitem}
\setlist[enumerate]{itemsep=1.5pt, topsep=2.5pt, parsep=0pt}

\definecolor{darkblue}{rgb}{0, 0, 0.5}
\hypersetup{colorlinks=true, citecolor=darkblue, linkcolor=darkblue, urlcolor=darkblue}

\newtheorem{proposition}{Proposition}

\theoremstyle{definition}
\newtheorem{definition}{Definition}
\newtheorem{hypothesis}{Hypothesis}
\theoremstyle{remark}
\newtheorem{remark}{Remark}

\newcommand{\R}{\mathbb{R}}
\newcommand{\dist}{\operatorname{dist}}
\newcommand{\F}{\mathcal{F}}

\newcommand{\res}{r}
\newcommand{\grad}{\nabla}
\DeclareMathOperator*{\argmin}{arg\,min}
\newcommand{\norm}[1]{\lVert #1 \rVert}
\newcommand{\beff}{b^{\mathrm{eff}}}
\newcommand{\pdist}{\widetilde d}

\newcommand{\ablDusable}{41.1}

\newcommand{\ablFusable}{90.0}

\newcommand{\ablFCusable}{20.0}

\newcommand{\ablFCVusable}{21.1}

\newcommand{\ablGateCoverage}{88.2}
\newcommand{\ablGateN}{34}
\newcommand{\ablToolEvents}{0}

\newcommand{\ablFCprecision}{62.2}
\newcommand{\ablFCVreturnedN}{57}

\newcommand{\ablGateCoveredN}{30}

\newcommand{\ablSDRtightN}{18}
\newcommand{\ablSDRtightU}{18}
\newcommand{\ablSDRlooseN}{34}
\newcommand{\ablSDRlooseFeas}{0}
\newcommand{\ablSCAN}{38}
\newcommand{\ablSCAfeasN}{38}
\newcommand{\ablSCAusableN}{0}

\newcommand{\ablStratMixedPids}{21}
\newcommand{\ablPairedFwins}{23}
\newcommand{\ablPairedDwins}{3}
\newcommand{\ablPairedTies}{4}
\newcommand{\ablSignP}{8.8\times 10^{-5}}
\newcommand{\ablFinitRandom}{73.0}

\newcommand{\ablSolverTotalF}{0.1}
\newcommand{\ablOracleCertN}{12}

\newcommand{\ablBeatOracleN}{5}

\newcommand{\entCalls}{120}
\newcommand{\entPairs}{60}
\newcommand{\entHPre}{1.092}
\newcommand{\entHPost}{0.679}
\newcommand{\entHDelta}{0.413}
\newcommand{\entHCILo}{0.306}
\newcommand{\entHCIHi}{0.523}
\newcommand{\entUMPre}{32.5\%}
\newcommand{\entUMPost}{38.9\%}
\newcommand{\entUMDelta}{6.3}
\newcommand{\entUMCILo}{0.6}
\newcommand{\entUMCIHi}{12.8}
\newcommand{\entFMPre}{71.1\%}
\newcommand{\entFMPost}{80.9\%}
\newcommand{\entFMDelta}{9.8}
\newcommand{\entFMCILo}{3.7}
\newcommand{\entFMCIHi}{16.1}
\newcommand{\entSelUsePre}{18.3\%}
\newcommand{\entSelUsePost}{36.7\%}
\newcommand{\entSelUseDelta}{18.3}
\newcommand{\entSelUseCILo}{8.3}
\newcommand{\entSelUseCIHi}{30.0}

\newcommand{\entChanged}{16}

\newcommand{\entChangedGood}{11}
\newcommand{\entChangedNeutral}{5}
\newcommand{\entChangedBad}{0}

\newcommand{\entUsefulReopen}{6}
\newcommand{\entOverContract}{28}

\newcommand{\epEffectiveCalls}{72}
\newcommand{\epActualRequests}{78}
\newcommand{\epHZero}{1.301}
\newcommand{\epHLate}{0.512}
\newcommand{\epHDelta}{0.789}
\newcommand{\epHCILo}{0.172}
\newcommand{\epHCIHi}{1.482}
\newcommand{\epMZero}{30.6\%}
\newcommand{\epMLate}{2.8\%}
\newcommand{\epMDeltaPP}{-27.8}
\newcommand{\epMCILoPP}{-61.1}
\newcommand{\epMCIHiPP}{0.0}
\newcommand{\epFZero}{33.3\%}
\newcommand{\epFLate}{86.1\%}

\newcommand{\egCalls}{24}
\newcommand{\egNewCalls}{12}
\newcommand{\egHDirect}{0.629}
\newcommand{\egHEndo}{0.629}
\newcommand{\egDirectUse}{25.0\%}
\newcommand{\egEndoUse}{8.3\%}
\newcommand{\egDirectLoose}{33.3\%}
\newcommand{\egEndoLoose}{8.3\%}
\newcommand{\egStateCalls}{10}
\newcommand{\egSoftStates}{16}
\newcommand{\egHardStates}{9}
\newcommand{\egRevisedStates}{1}
\newcommand{\egCheckClaims}{8}
\newcommand{\egCheckCorrect}{1}
\newcommand{\egWallDirect}{184.0}
\newcommand{\egWallEndo}{222.5}

\newcommand{\egHWins}{2}
\newcommand{\egHLosses}{2}
\newcommand{\egHTies}{2}

\title{Constrained Path Reasoning:\\
Measuring When Committed Stages Earn Their Cost}

\author{Honglin Li\\
ShanghaiTech University\\
\texttt{lihl2025@shanghaitech.edu.cn}}

\begin{document}

\ifcolmsubmission
\linenumbers
\fi

\maketitle

\begin{abstract}
When does a committed intermediate stage in an LLM reasoning pipeline earn
its cost? \emph{Constrained Path Reasoning} (CPR) pairs a source-aware path
hypothesis with stage-level accounting. Search generates provisional states;
trusted or validated invariants can constrain hard, while other proposals
remain soft and revisable. CPR predicts that task-compatible commitments can
factor transitions, concentrate candidate mass, induce regularity, and expose
feedback when their gains exceed propagated error and execution cost. The
formalism covers discrete commitments and continuous flows and measures
effective branching, endpoint concentration, and cost per usable output.
Across $1{,}180$ generated QCQPs and $40$ engineered degenerate polynomial
instances ($2{,}140$ endpoints), residual triage recovers $63.0\%$ of
repair-all's additional feasible yield with $17.7\%$ of its attempts.
Fixed-LLM accounting ($270$ unique calls shared across nested arms) finds
usable yield of $\ablDusable\%$ direct, $\ablFusable\%$ after formalization
and deterministic execution, $\ablFCusable\%$ after one-shot
convexification, and $\ablFCVusable\%$ for the full path. In \entCalls{}
paired-condition calls, a two-action rollback rule reaches $90\%$ usable
yield versus \entSelUsePost{} for the feedback-conditioned selector. Two
endpoint probes separate source from validation: a \epEffectiveCalls-output
cross-trajectory transplant reduces entropy and acceptable mass; a
\egCalls-output same-call self-proposal pilot gives unchanged two-repeat
collision entropy, \egDirectUse{} versus \egEndoUse{} usable yield, and
\egCheckCorrect{}/\egCheckClaims{} deterministically confirmed endpoint
checks. Model-generated states supply hypotheses; trusted execution earns
constraint strength.
\end{abstract}

\section{Introduction}
\label{sec:intro}

Explicit intermediate reasoning improves many LLM tasks
\citep{wei2022chain,kojima2022large,wang2023selfconsistency,snell2024scaling},
while excessive reasoning can waste compute or reduce accuracy
\citep{chen2024not,sui2025stop}. We study a complementary design question:
how do committed intermediate states change the search, execution, and
verification available to later steps?

\paragraph{Constrained Path Reasoning.} We view CoT as endogenous constrained
path construction. At each step, the model's heuristic search proposes a
local continuation or intermediate artifact, selects one, and conditions
subsequent search on it. Trusted invariants---for example, a physical law or
a verified formal contract---may restrict the path hard. Model-generated
subgoals, decompositions, and surrogate assumptions remain provisional until
validated, retaining revision and rollback. CPR hypothesizes that
task-compatible constraints can factor a long transition into shorter
operators, concentrate candidate mass, induce regularity, and expose
correction signals. The same account covers single-path decoding and
explicit multi-branch search
\citep{yao2023tree,besta2024graph,hao2023reasoning}.

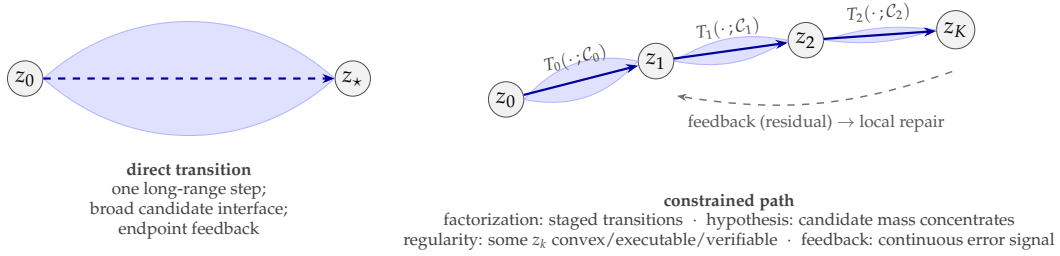
\begin{figure}[t]
\centering
\resizebox{\linewidth}{!}{%
\begin{tikzpicture}[
  state/.style={circle, draw=black!70, fill=black!5, inner sep=1.5pt, minimum size=14pt, font=\small},
  fan/.style={fill=blue!14, draw=blue!35, line width=0.3pt, opacity=0.85},
  lab/.style={font=\scriptsize, text=black!75, align=center},
  pstep/.style={-{Stealth[length=5pt]}, line width=0.9pt, draw=blue!60!black},
]
\node[state] (z0) at (0,0) {$z_0$};
\node[state] (zs) at (4.7,0) {$z_\star$};
\path[fan] (z0.east) to[bend left=42] (zs.west) to[bend left=42] cycle;
\draw[pstep, dashed] (z0) -- (zs);
\node[lab] at (2.35,-1.75) {\textbf{direct transition}\\ one long-range step;\\ broad candidate interface;\\ endpoint feedback};
\begin{scope}[shift={(6.9,0)}]
\node[state] (y0) at (0,-0.3) {$z_0$};
\node[state] (y1) at (2.15,0.25) {$z_1$};
\node[state] (y2) at (4.3,0.55) {$z_2$};
\node[state] (y3) at (6.45,0.7) {$z_K$};
\path[fan] (y0.east) to[bend left=30] (y1.west) to[bend left=30] cycle;
\path[fan] (y1.east) to[bend left=21] (y2.west) to[bend left=21] cycle;
\path[fan] (y2.east) to[bend left=12] (y3.west) to[bend left=12] cycle;
\draw[pstep] (y0) -- node[above=1.5pt, sloped, font=\scriptsize, text=black!75]{$T_0(\cdot\,;\mathcal C_0)$} (y1);
\draw[pstep] (y1) -- node[above=1.5pt, sloped, font=\scriptsize, text=black!75]{$T_1(\cdot\,;\mathcal C_1)$} (y2);
\draw[pstep] (y2) -- node[above=1.5pt, sloped, font=\scriptsize, text=black!75]{$T_2(\cdot\,;\mathcal C_2)$} (y3);
\draw[-{Stealth[length=4pt]}, line width=0.6pt, draw=black!55, dashed]
  ($(y3)+(-0.05,-0.6)$) to[bend left=14] node[below=1pt, font=\scriptsize, text=black!65]{feedback (residual) $\rightarrow$ local repair} ($(y1)+(0.3,-0.5)$);
\node[lab] at (3.2,-2.05) {\textbf{constrained path}\\ factorization: staged transitions \;$\cdot$\; hypothesis: candidate mass concentrates\\ regularity: some $z_k$ convex/executable/verifiable \;$\cdot$\; feedback: continuous error signal};
\end{scope}
\end{tikzpicture}}
\caption{Constrained Path Reasoning. \textbf{Left:} direct prediction crosses
a broad candidate interface and receives endpoint feedback. \textbf{Right:}
trajectory-generated states or trusted invariants factor the transition, may
concentrate candidate mass, induce regularity, and expose correction signals.
Provisional states retain rollback. The fans visualize the
candidate-level hypothesis; arrows may be discrete operators or checkpointed
continuous flows, and experiments use fixed action and endpoint interfaces.}
\label{fig:cpr}
\end{figure}

The theoretical claims become testable at committed-stage interfaces. CPR
records artifact validity, downstream utility, operational cost, induced
regularity, feedback, candidate concentration, constraint provenance, and
whether an unverified state remains corrigible. Reasoning efficiency includes
tokens, executor time, repair effort, and discarded outputs, so each stage is
also priced by cost per usable solution. Section~\ref{sec:cpr-mech} develops
the ambiguity--error--cost hypothesis behind this accounting
(Fig.~\ref{fig:cpr}).

\paragraph{A quantitative case study: automated convexification.} We ground
CPR in a domain with executable stage interfaces and numerical feedback:
LLM-assisted convexification of non-convex optimization problems, as
instantiated by the recent NC2C system \citep{peng2026nc2c} and related
LLM-for-optimization pipelines \citep{ahmaditeshnizi2024optimus,xiao2024chain}.
The controlled path uses literal coefficient transcription, deterministic
execution from a standardized feasible initialization, and one committed SDR
or OSM surrogate transition. The convex surrogate supports numerical solving
and lower-bound diagnostics where available; its endpoint exposes the
\emph{constraint residual} against the original constraints. Classical error-bound
theory \citep{hoffman1952approximate,pang1997error} maps this residual to a
distance bound under standard regularity. Building on NC2C's validation and
FDC repair interface, we operationalize the relation as residual-guided
endpoint triage. This case study prices the adoption and execution of explicit
artifacts. A same-call pilot additionally elicits autonomous state proposals
from the natural-language problem alone.

\paragraph{Contributions.}
\begin{enumerate}
\item \textbf{Stage-level accounting protocol} (\S\ref{sec:cpr}): for
committed executable interfaces, we combine shared-call nested ablations,
stage-local checks, constraint provenance/corrigibility, executor-aware cost,
and utility per usable endpoint to localize observed changes.
\item \textbf{Source-aware CPR hypothesis} (\S\ref{sec:cpr}): verified
invariants can constrain hard, while trajectory-generated intermediate states
remain defeasible; discrete stages and continuous flows are organized through
factorization, concentration, regularity, feedback, and propagated error.
\item \textbf{Quantitative case study}
(\S\S\ref{sec:case}--\ref{sec:exp-ent}): a residual verifier, fixed-LLM
evaluation, and endpoint probe expose relaxation leakage, over-contraction,
initialization sensitivity, a feedback-policy gap, and the failure of a
cross-trajectory surrogate-state transplant; a same-call self-proposal pilot
separates proposal generation from executable validation.
\end{enumerate}

\section{Constrained path reasoning}
\label{sec:cpr}

\subsection{Autoregressive reasoning as sequential heuristic search}
\label{sec:cpr-search}

Let $x$ be the input and $s_k$ the $k$-th committed reasoning state: a token
prefix together with any explicit artifact it encodes, such as a subgoal,
formal program, or tool result. The local successor distribution
$q_\theta(z\mid s_k,x)$ acts as a learned heuristic over plausible
continuations. A decoder $D$ then commits one continuation and updates the
state:
\begin{equation}
\label{eq:decode}
\hat z_{k+1}=D\!\left(q_\theta(\cdot\mid s_k,x)\right),
\qquad s_{k+1}=U(s_k,\hat z_{k+1}).
\end{equation}
Thus each step follows propose/score $\to$ select $\to$ condition the next
search. Here a \emph{commitment} is a state selected strongly enough to
condition later search while remaining eligible for revision. Single-path
decoding performs local scoring implicitly; beam search,
Tree-of-Thoughts \citep{yao2023tree}, graph variants
\citep{besta2024graph}, and MCTS-style planning \citep{hao2023reasoning}
make candidate search explicit. CPR studies the selected path produced in
either regime.

\paragraph{Constraint source and strength.}
The path constraint is generally produced during search. We decompose it as
\begin{equation}
\label{eq:constraint-source}
\widehat{\mathcal C}_k\sim
G_\theta(\cdot\mid s_k,x),\qquad
\mathcal C_k=\mathcal I_k\cup\widehat{\mathcal C}_k ,
\end{equation}
where $\mathcal I_k$ contains trusted exogenous invariants and
$\widehat{\mathcal C}_k$ is an endogenous proposal---a subgoal, decomposition,
formal artifact, or local hypothesis---generated from the current trajectory.
Here $a_k(z;\widehat{\mathcal C}_k)$ scores proposal compatibility, and
$\lambda_k$ is its validation strength.
\begin{equation}
\label{eq:constrained}
q_\theta^{\mathcal C_k}(z\mid s_k,x)
\propto q_\theta(z\mid s_k,x)\,
\mathbf 1\!\left[z\in\Omega_k(s_k;\mathcal I_k)\right]\,
\exp\!\left\{\lambda_k a_k(z;\widehat{\mathcal C}_k)\right\}.
\end{equation}
Trusted invariants, such as a known physical law, receive hard support
restriction. A model-generated proposal receives finite weight and retains
revision, rollback, or escape until a verifier certifies it; certification
strength comes from a trusted derivation or executor contract. A certified
proposal can approach the hard-constraint limit. We call the deployed
constraint \emph{support-preserving} when it retains positive mass on
continuations extendable to a correct answer. It is \emph{task-compatible} at
tolerance $\tau$ when
\[
\Pr(\mathrm{success}\mid s_k,\mathcal C_k)
\ge \Pr(\mathrm{success}\mid s_k)-\tau.
\]
For a fixed discrete candidate representation, the candidate-level effective
branching factor is
\begin{equation}
\label{eq:beff}
\beff_k=\exp\!\left(
H\!\left[q_\theta^{\mathcal C_k}(\cdot\mid s_k,x)\right]\right).
\end{equation}
Continuous states require a fixed quantization or a separately specified
description-length scheme. CPR predicts that useful paths reduce accumulated
candidate ambiguity, or the search work it induces, while preserving task
utility.

At the operator level, let $z_0$ encode the problem and $z_\star$ an
acceptable endpoint. A structured path is
\begin{equation}
\label{eq:path}
z_0\to z_1\to\dots\to z_K=z_\star,
\qquad Z_{k+1}\sim P_k(\cdot\mid Z_k;\mathcal C_k),
\end{equation}
where $P_k$ wraps an LLM call, symbolic transformation, solver, or verifier.
The kernels localize transcription, approximation, and numerical error to the
stage that creates it. Where numeric states embed in a normed space, we write
$z_{k+1}=T_k(z_k;\mathcal C_k)+\xi_k$; the kernel form also covers symbolic
and heterogeneous state spaces.

\paragraph{Continuous constrained paths.} Let
$\mathcal M_t^{\mathrm{inv}}=\{z:\phi(z,t)=0\}$ encode a trusted prescribed
or independently certified manifold. A continuous CPR segment satisfies
\[
\dot z_t=f_\theta(z_t,t,x)+u_t,\qquad
\phi(z_t,t)=0,\qquad
D_z\phi\,\dot z_t+\partial_t\phi=0.
\]
Here $f_\theta$ may combine a trusted prior ODE $f_0$ with a learned
tangential residual, and $u_t$ enforces the invariant. An uncertified learned
manifold enters as a finite-weight potential or confidence tube, preserving
the same corrigibility as $\widehat{\mathcal C}_k$; certification promotes it
to $\mathcal M_t^{\mathrm{inv}}$. Discretization recovers
Eq.~\eqref{eq:path}; hybrid paths alternate continuous segments with symbolic
commitments. Fixed probes and probability flow give endpoint, entropy-rate,
and reachable-volume diagnostics (App.~\ref{app:continuous}).

\subsection{Four mechanisms and an ambiguity--error--cost hypothesis}
\label{sec:cpr-mech}

The process in \S\ref{sec:cpr-search} motivates four mechanisms:
\begin{enumerate}
\item \textbf{Transition factorization.} Intermediate states replace one
endpoint transition by staged operators. CPR hypothesizes that semantically
shorter operators can reduce per-step function complexity and
out-of-distribution span, consistent with locality accounts of CoT
\citep{prystawski2023why} and expressivity results for intermediate tokens
\citep{merrill2024expressive,li2024chain}.
\item \textbf{State-space contraction.} A useful constraint can concentrate
successor mass. At a fixed candidate representation, its entropy change is
\begin{equation}
\label{eq:contraction}
\Delta H_k:=H[q_\theta(\cdot\mid s_k,x)]
-H[q_\theta^{\mathcal C_k}(\cdot\mid s_k,x)].
\end{equation}
We call the step entropy-contracting when $\Delta H_k>0$ and treat its sign as
an empirical property. Section~\ref{sec:exp-ent} instantiates the diagnostic
at a fixed A--D action menu using schema-elicited allocations; logged
candidate probabilities would instantiate it at the model-successor level.
Measuring the model-level endpoint distribution directly is a planned
follow-up: it requires token-level probability access or a substantially
larger repeated-call budget. The present study instead prices a cost-bounded
interface diagnostic.
\item \textbf{Regularity induction.} Committed states can supply convexity,
smoothness, executability, or verifiability, making the next operator
well-posed and its output inspectable.
\item \textbf{Feedback and correction.} Intermediate states can expose
\emph{continuous} error signals (a residual, a failing test) so local repair
replaces end-to-end regeneration; process supervision exploits the same
observability in training \citep{lightman2024lets,uesato2022solving},
self-refinement at inference \citep{madaan2023selfrefine,shinn2023reflexion}.
\end{enumerate}

\paragraph{Error propagation.} For numeric or metrically embedded stages,
suppose
\begin{equation}
\label{eq:error}
\norm{e_{k+1}}\le L_k\norm{e_k}+\eta_k,
\qquad
\norm{e_K}\le
\left(\prod_{i=0}^{K-1}L_i\right)\norm{e_0}
+\sum_{j=0}^{K-1}\eta_j\prod_{i=j+1}^{K-1}L_i,
\end{equation}
where $L_k$ is local sensitivity and $\eta_k$ is error injected at stage $k$.
The expanded bound assigns every injected error its downstream sensitivity.
Together with \eqref{eq:contraction}, it yields CPR's design prediction: a
path is valuable when ambiguity reduction and regularity gains outweigh
propagated error, task-utility loss, and execution cost. Over-contraction,
relaxation leakage, and large suffix sensitivities are corresponding failure
modes. For an unverified endogenous state, finite constraint strength and an
available revision or rollback action preserve the search's corrigibility.

\begin{hypothesis}[Pathwise endpoint concentration]
\label{hyp:ambiguity}
Fix a discrete (or fixed-quantized) endpoint candidate variable $Y$ with
representation $\mathcal Y$, an
endpoint probe, and an acceptance set
$\mathcal Y_{\mathrm{acc}}(x)\subseteq\mathcal Y$. Let
$Q_\theta^{\mathrm{end}}(\cdot\mid S_k,x)$ be the model-implied distribution
returned by that same probe from committed state $S_k$, and define
\begin{equation}
\label{eq:hypothesis}
\mathcal A_k :=
\mathbb E_{x,S_k}\!\left[
H\!\left(Q_\theta^{\mathrm{end}}(\cdot\mid S_k,x)\right)\right].
\end{equation}
The expectation follows the task distribution and the path policy. Its
provisional states are generated within the sampled trajectory by
\eqref{eq:constraint-source}; hard state components come from trusted or
verified invariants.
For task--model pairs where structured reasoning is beneficial, CPR
conjectures a strict net concentration before endpoint generation,
\begin{equation}
\label{eq:endpoint-concentration}
\mathcal A_{K-1}<\mathcal A_0,
\qquad
\mathbb E_{x,S_{K-1}}\!\left[
Q_\theta^{\mathrm{end}}
  (\mathcal Y_{\mathrm{acc}}(x)\mid S_{K-1},x)\right]
\ge
\mathbb E_{x,S_0}\!\left[
Q_\theta^{\mathrm{end}}
  (\mathcal Y_{\mathrm{acc}}(x)\mid S_0,x)\right]-\tau .
\end{equation}
\end{hypothesis}
This compares the same endpoint variable along the path. Discrete stage
contributions are
$\Delta_k^{\mathrm{end}}:=\mathcal A_k-\mathcal A_{k+1}$. On a continuous
segment, write $S_t=(z_t,t)$, define $\mathcal A(t)$ by
\eqref{eq:hypothesis}, and use the concentration rate
$\chi(t)=-d\mathcal A(t)/dt$, so
$\int_0^T\chi(t)dt=\mathcal A(0)-\mathcal A(T)$. Local quantities may have
either sign: exploration, rollback, or reflection can expand the endpoint
distribution temporarily. The acceptance-mass condition distinguishes useful
concentration from confidence in an unusable endpoint. Local successor
entropy and fixed endpoint probes measure the immediate candidate interface
and final-task uncertainty, respectively, for discrete or checkpointed
continuous paths.

\paragraph{Operational path accounting.} For a path design $\pi$, let
$U_\pi$ indicate a usable endpoint and let $C_\pi^{(m)}$ be a fixed cost
metric $m$---tokens, wall-clock, continuous-flow function evaluations, or a
declared price for LLM inference, solving, verification, and repair. The target is
\begin{equation}
\label{eq:costuse}
\mathcal C_{\mathrm{use}}^{(m)}(\pi)
:= \frac{\mathbb E[C_\pi^{(m)}]}{\Pr(U_\pi=1)},
\end{equation}
estimated by total cost divided by usable outputs. A committed stage earns
its cost when an ablation shows lower $\mathcal C_{\mathrm{use}}^{(m)}$, greater
yield under a fixed budget, or higher returned-answer precision at an
acceptable coverage. Action concentration and effective action count are
reported only when their candidate interface, prompt, and state encoding are
fixed; the four-action allocation supplies the executable local measure here.
The audit additionally records whether each constraint is trusted,
verified, or provisional, and whether the suffix exposes revision or
rollback.

\paragraph{Accounting protocol.} Given nested path packages
$\pi_0,\ldots,\pi_K$, local contracts, fixed executors, a usable-endpoint
criterion, and a cost vector, the protocol reuses upstream calls, checks each
new artifact locally, executes endpoints against common ground truth, and
reports changes in yield, coverage/precision,
$\mathcal C_{\mathrm{use}}^{(m)}$, and localized errors. The resulting audit
is indexed by the evaluated task, model, executor, and cost vector.

\paragraph{Relation to existing views.} That reasoning is search is classical
\citep{newell1976computer}, and recent work makes LLM reasoning search-like
explicitly \citep{yao2023tree,hao2023reasoning,gandhi2024stream}; constrained
decoding imposes structural constraints at the token level---incremental
parsing \citep{scholak2021picard}, predicate-logic constraints
\citep{lu2021neurologic}, knowledge-graph faithfulness
\citep{luo2024graph}---and Constraints-of-Thought
\citep{alrashedy2025constraints} turns reasoning steps into executable
constraints inside a planner. CPR links these views through a common claim:
task-compatible commitments can factor transitions, concentrate successor
mass, induce regularity, and expose feedback. Its accounting layer then
measures task utility, cost, provenance, corrigibility, artifact validity, and
stage-localized error at the same stage boundary. Committed states may also be non-textual
\citep{hao2024coconut}.

\section{Case study: committed stages in a convexification pipeline}
\label{sec:case}

\begin{figure}[t]
\centering
\resizebox{0.92\linewidth}{!}{%
\begin{tikzpicture}[
  st/.style={draw=black!65, rounded corners=2pt, fill=black!4, minimum height=26pt,
             inner sep=4.5pt, font=\small, align=center},
  op/.style={font=\scriptsize, text=black!60, align=center},
  ar/.style={-{Stealth[length=5pt]}, line width=0.8pt, draw=blue!55!black},
]
\node[st] (d) at (0,0) {NL problem\\ $z_0$};
\node[st] (p) at (3.2,0) {formal\\ program $\mathcal P$};
\node[st] (n) at (6.6,0) {non-convex\\ structure};
\node[st] (pc) at (10.2,0) {convex\\ surrogate $\mathcal P_c$};
\node[st] (xc) at (13.5,0) {solver\\ solution $x_c$};
\node[st, fill=blue!8] (tri) at (16.5,0) {residual\\ triage};
\draw[ar] (d) -- node[op, above=2pt]{LLM\\ formalize} (p);
\draw[ar] (p) -- node[op, above=2pt]{LLM\\ detect} (n);
\draw[ar] (n) -- node[op, above=2pt]{LLM strategy\\ (SDR/SCA/\dots)} (pc);
\draw[ar] (pc) -- node[op, above=2pt]{convex\\ solver} (xc);
\draw[ar] (xc) -- node[op, above=2pt]{$r(x_c)$} (tri);
\node[op, below=7pt of pc] {\emph{regularity induction}\\[-1pt]
outer: $\F\subseteq C_{\mathrm{relax}}$, $r>0$ possible\\[-1pt]
inner: $C_{\mathrm{inner}}\subseteq\F$, $r=0$, conservative};
\node[op, below=7pt of tri] {\textsc{near-feasible} /\\[-1pt] \textsc{repair} / \textsc{reject}};
\node[op, below=7pt of p] {committed\\[-1pt] symbolic state};
\node[op, below=7pt of xc] {repair: bounded\\[-1pt] local correction};
\draw[-{Stealth[length=4pt]}, line width=0.6pt, draw=black!55, dashed]
  (tri.north) to[out=150, in=30] node[op, above=1pt]{reject $\Rightarrow$ re-convexify (new strategy)} (pc.north);
\draw[-{Stealth[length=4pt]}, line width=0.6pt, draw=black!55, dashed]
  ([xshift=-8pt]tri.south) to[out=-120, in=-60] ([xshift=8pt]xc.south);
\end{tikzpicture}}
\caption{Automated convexification as a constrained path with heterogeneous
operators. LLM steps commit symbolic states, the convex surrogate induces
regularity, and the endpoint residual drives triage and local correction.
Outer and inner approximations relate to $\F$ in opposite directions.}
\label{fig:case}
\end{figure}
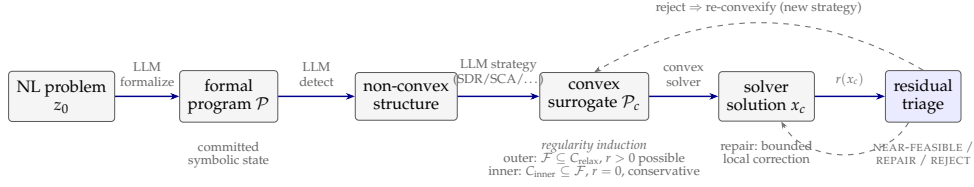

\paragraph{NC2C as a constrained path.} NC2C \citep{peng2026nc2c} maps a
natural-language non-convex optimization problem through formalization,
non-convexity detection, convexification, code execution, and feasibility
repair. Its heterogeneous states are $z_0=$ description, $z_1=\mathcal P$,
$z_2=$ detected structure, $z_3=\mathcal P_c$, $z_4=x_c$, and $z_5=$ triage
or repaired point (Fig.~\ref{fig:case}). LLM stages construct formal artifacts,
a deterministic solver executes $\mathcal P_c$, and a verifier/repair operator
checks the original constraints. The model-proposed surrogate $z_3$ is a
provisional path state: it supplies regularity, while its endpoint supplies
the continuous residual
\begin{equation}
\label{eq:res}
\res(x) \;=\; \max\Big\{\max_i \big(g_i(x)\big)_+,\; \max_j |h_j(x)|\Big\},
\end{equation}
the violation of the original constraints $g_i\le 0$, $h_j=0$ by the
surrogate's solution. Appendix~\ref{app:proof} specifies the residual
conventions and local equivalences used by the harness. Repair,
re-convexification, and rollback retain corrigibility after this proposal.

\paragraph{Geometry of the convexification step.} The two main
convexification families relate to $\F$ in opposite directions. An
\emph{outer} relaxation (SDR \citep{luo2010semidefinite}, McCormick envelopes
\citep{mccormick1976computability}) satisfies $\F\subseteq
C_{\mathrm{relax}}$: its optimum lower-bounds the original optimal value and
may have $\res>0$. An \emph{inner} approximation (SCA-type majorization
anchored at a feasible point with no slack, as in our harness
\citep{lipp2016variations}) satisfies $C_{\mathrm{inner}}\subseteq\F$: its
solutions have $\res=0$ by construction but can be conservative. Together
with the symbolic formalization and detection stages, these relations produce
the heterogeneous path in Fig.~\ref{fig:case}. For outer relaxations,
$\res(x_c)$ is the computable scalar that connects the surrogate endpoint to
the original constraints and turns binary validation into graded feedback.

\section{Residual-guided endpoint verification}
\label{sec:verify}

We now make the feedback channel quantitative. Throughout, $\F=\{x \in
\R^n : g_i(x)\le 0,\, h_j(x)=0\}$ is the original feasible set, assumed
non-empty, with $g_i,h_j \in C^1$ and $\dist(x,\F) := \inf_{y\in\F}
\norm{x-y}_2$. We assume a bounding constraint among the $g_i$ (in our instances the
smooth ball $\norm{x}^2 - R^2 \le 0$, which also enters the residual), so
$\F$ and the residual tubes below are compact.

\subsection{Classical error bounds, operationalized}

\begin{definition}[H\"olderian residual bound]
\label{def:eb}
The constraint system admits a residual bound with exponent $\theta \in (0,1]$
and modulus $\kappa>0$ on the tube $U_{\delta_0} = \{x \in \R^n : \res(x)
\le \delta_0\}$ if
\begin{equation}
\label{eq:EB}
\dist(x, \F) \;\le\; \kappa\, \res(x)^{\theta}
\qquad \text{for all } x \in U_{\delta_0}.
\tag{EB$_\theta$}
\end{equation}
\end{definition}

Bounds of this form have a mature literature: Hoffman's bound for linear
systems \citep{hoffman1952approximate}, error bounds under constraint
qualifications
\citep{pang1997error,rockafellar1998variational,ngai2008error}, H\"olderian
bounds under degeneracy \citep{luo1994extension,bolte2017error}, quadratic
growth in convex settings \citep{drusvyatskiy2018error}. We use these
classical bounds as the decision layer of the convexification path: they map
an endpoint residual to an acceptance scale or a repair budget.

\begin{proposition}[Residual triage; operationalization of classical bounds]
\label{prop:triage}
Assume \eqref{eq:EB} holds on $U_{\delta_0}$. Let $0 < \varepsilon \le
\beta$ be an acceptance accuracy and a repair radius budget (so the two
thresholds below are ordered), and let $x_c$ satisfy
$\res(x_c) \le \delta_0$. Then:
\begin{enumerate}
\item[(i)] if $\res(x_c) \le (\varepsilon/\kappa)^{1/\theta}$, some
$x^\star \in \F$ has $\norm{x_c - x^\star} \le \varepsilon$
(\textsc{near-feasible}: a proximity certificate);
\item[(ii)] if $\res(x_c) \le (\beta/\kappa)^{1/\theta}$, then
$\dist(x_c,\F) \le \beta$, so a feasible point lies within the repair budget
(\textsc{repair});
\item[(iii)] if $\res(x_c)>(\beta/\kappa)^{1/\theta}$, the bound does not
certify the existence of a feasible point within radius $\beta$
(\textsc{reject}).
\end{enumerate}
At every feasible point satisfying MFCQ, a local linear residual bound holds
\citep{robinson1976stability,rockafellar1998variational,pang1997error}. If
MFCQ holds throughout compact $\F$, finitely many local moduli give a uniform
linear bound on a neighborhood of $\F$ (and on a fixed compact residual tube
as detailed in App.~\ref{app:proof}). If instead
$\res(x)\ge\mu\dist(x,\F)^2$, \eqref{eq:EB} holds with
$\theta=\tfrac12$ and $\kappa=\mu^{-1/2}$.
\end{proposition}

The tiers certify proximity and repair radius; endpoint yield additionally
depends on the repair algorithm. We therefore count numerically feasible
outputs and evaluate a fixed bounded correction directly.

\subsection{Analytical certificate vs.\ calibrated gate}
\label{sec:gate}

For a verified constraint family, Proposition~\ref{prop:triage} supplies
analytic $(\kappa,\theta)$. For a calibrated family, let
$\pdist(x_c,\F):=\min_{\ell\le L}\norm{x_c-\tilde x_\ell}_2$, where each
$\tilde x_\ell\in\F$ is a feasible return from a fixed multi-start projection
protocol; hence $\pdist(x_c,\F)\ge\dist(x_c,\F)$. We fit $\hat\theta$ by
unconstrained log--log regression of $\pdist(x_c,\F)$ on $\res(x_c)$ and set
$\hat\kappa$ to the 95th percentile
of $\pdist/\res^{\hat\theta}$. We report fitted slopes above one as finite-range
empirical power-law exponents and summarize their decision semantics by
held-out proxy coverage within the calibration residual range. Verified
regularity provides analytical certificates; the fitted gate provides the
reported empirical coverage diagnostic.

\paragraph{The three-tier verifier.}
\label{sec:algo}
The verifier computes $\res(x_c)$ and returns
\textsc{near-feasible} if $\res(x_c) \le \delta_1 :=
(\varepsilon/\hat\kappa)^{1/\hat\theta}$; else \textsc{repair}---invoke a
bounded local correction with radius budget $\beta$---if $\res(x_c) \le
\delta_2 := (\beta/\hat\kappa)^{1/\hat\theta}$; else \textsc{reject} (switch
strategy or re-convexify). Analytical thresholds are capped by the bound's
validity radius; calibrated thresholds operate within empirical support. This
maps a geometric repair budget to a residual cutoff and spends correction
only on endpoints predicted to lie within reach. For deployment, a
\textsc{near-feasible} point that still exceeds the numerical tolerance also
receives correction; at our operating point these 16 points plus 37
\textsc{repair} points produce the 53 attempts in Table~\ref{tab:triage}.
The analytical budget $\beta$ is Euclidean distance; the harness adopts the
stricter cumulative repair-path-length cap at the same value. All reported
yields require numerical feasibility after correction.

\section{Experiments I: case-study validation}
\label{sec:exp-cs}

We first test the quantitative core on controlled non-convex problems
(details, seeds, and per-instance raw data: App.~\ref{app:cs-details} and
the supplementary material).

\paragraph{Setup.} We generate feasible non-convex QCQPs
$\min_x c^\top x$ subject to $x^\top A_kx+b_k^\top x+d_k\le0$ and
$\norm{x}\le R$, using $m\approx3$--$4n$ indefinite constraints. SDR is
solved with CVXPY \citep{diamond2016cvxpy}. The fixed multi-start SLSQP
protocol supplies $\pdist(x_c,\F)$, the feasible projection proxy defined in
\S\ref{sec:gate}; bounded projected-subgradient descent records path length
to first feasibility. All instances are generated and executed with standard
numerical solvers and fixed seeds.

\begin{figure}[t]
\centering
\includegraphics[width=\linewidth]{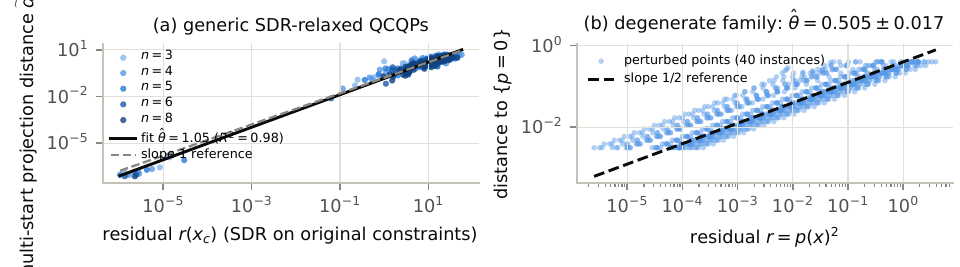}
\caption{Residual--projection-distance scaling on generic SDR-relaxed QCQPs
and engineered degenerate polynomial instances. \textbf{(a)}
Generic QCQPs: $\pdist(x_c,\F) \approx 0.145\, r^{1.05}$, where
$\pdist\ge\dist$ is the multi-start projection proxy,
$R^2=0.98$ ($n=325$ instances, colored by dimension). \textbf{(b)}
Engineered degenerate family (vanishing active gradients): within-instance
exponent $\hat\theta = 0.505\pm0.017$, the quadratic-growth regime.}
\label{fig:rd}
\end{figure}

\paragraph{Residual--projection-distance relation.} On $325$ SDR-relaxed
endpoints with $\res>0$ (Fig.~\ref{fig:rd}a), unconstrained log--log
regression gives $\pdist\approx0.145\,\res^{1.046}$ (95\% bootstrap CI
$[1.019,1.062]$, $R^2=0.977$) across roughly eight decades. The fitted slope
is a pooled finite-range calibration coefficient; the H\"older exponent in
Proposition~\ref{prop:triage} is system-local. An independent $80$-problem
check gives slope $1.024$ with $10$, $50$, or $100$ projection starts; the
densest protocol improves $11/80$ proxies and leaves their median ratio at
$1.00$. With $\hat\kappa=0.350$, this stable near-unit fit defines the
empirical gate; scaling remains stable through $n=12$
(App.~\ref{app:cs-details}). On $40$ engineered systems
$g(x)=p(x)^2\le0$, the within-instance exponent is
$0.505\pm0.017$ (Fig.~\ref{fig:rd}b), realizing the quadratic-growth regime.

\paragraph{Triage under a common repair operator.} On $360$ held-out
instances at $\beta=0.25R$, the gate reaches $28.3\%$ feasible yield with
$53$ repairs, versus $35.0\%$ with $299$ for repair-all
(Table~\ref{tab:triage}). It recovers
$(28.3-16.9)/(35.0-16.9)=63.0\%$ of repair-all's additional yield using
$53/299=17.7\%$ of its attempts, with $77.4\%$ repair precision. The
calibrated relation covers $342/360=95.0\%$ of held-out projection proxies
(Wilson CI $[92.2,96.8]\%$; residual--proxy Spearman $0.932$). At this budget,
the gate and residual ranking select the same $53$ endpoints; calibration
adds a geometric budget interpretation and held-out coverage. The cutoff
sweep appears in App.~\ref{app:cs-details}, Fig.~\ref{fig:frontier}. With repair price $c_r$ and
unresolved-endpoint retry price $c_v$, the observed counts give
$C_{\rm gate}=53c_r+258c_v$ and $C_{\rm all}=299c_r+234c_v$, so gating is
cheaper exactly when $c_r/c_v>0.098$ (derivation in
App.~\ref{app:cs-details}).

\begin{table}[t]
\begin{center}
\small
\begin{tabular}{lccc}
\toprule
Endpoint policy & usable yield & repair attempts & repair precision \\
\midrule
Binary accept-only (NC2C Eq.~14) & $16.9\%$ & $0$ & -- \\
Repair-all (budget $\beta=0.25R$) & $35.0\%$ & $299$ & $21.7\%$ \\
Residual-gated triage (ours) & $28.3\%$ & $53$ & $77.4\%$ \\
\bottomrule
\end{tabular}
\end{center}
\caption{Endpoint verification on $360$ held-out SDR-relaxed instances.
``Usable yield'' is the fraction of instances ending numerically feasible;
``repair attempts'' are bounded local-correction runs; precision is successful
repairs per attempt.}
\label{tab:triage}
\end{table}

\begin{figure}[t]
\centering
\includegraphics[width=\linewidth]{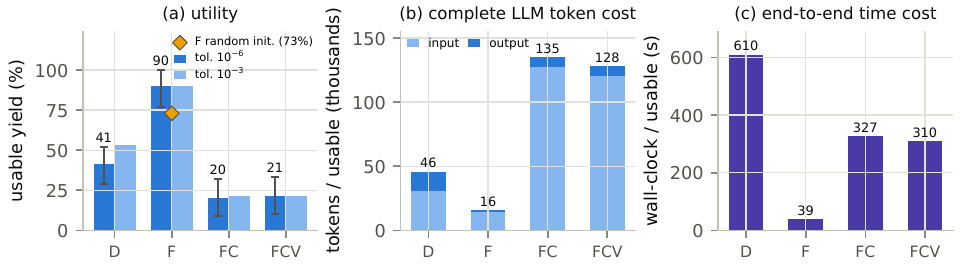}
\caption{Stage-package pricing. \textbf{(a)} Usable yield at two feasibility
tolerances; strict-yield error bars are problem-level bootstrap intervals, and
the diamond marks F with seeded random initialization. \textbf{(b)} Complete
LLM input and output tokens per usable solution. \textbf{(c)} LLM plus
executor wall-clock per usable solution.}
\label{fig:abl}
\end{figure}

\section{Experiments II: path ablation with an LLM}
\label{sec:exp-abl}

The case-study experiments isolate the feedback mechanism. We next run a
controlled LLM ablation to price the successive stage packages against direct
prediction.

\paragraph{Design.} We generate $30$ natural-language QCQPs ($n\in\{2,3,4\}$,
$m=3n$) whose rounded coefficients appear verbatim, defining a controlled
literal-transcription regime. Each problem receives three samples per arm
from fixed \texttt{gpt-5.5} settings (medium effort, schema-enforced JSON):
\textbf{D} returns a numeric solution directly; \textbf{F} commits a formal
program executed by SLSQP; \textbf{FC} additionally selects SDR or one-shot
inner majorization (OSM); and \textbf{FCV} adds residual triage and bounded
repair. D--F compares formal representation plus deterministic execution with
direct generation under literal coefficient transcription and a standardized
strictly feasible initialization $x=0$. OSM instantiates one committed
SCA-style surrogate transition anchored at that initialization. A result is usable when it is
$10^{-6}$-feasible and within $0.05R$ of the best-known 26-start oracle.
Appendix~\ref{app:abl-details} provides prompts, schemas, transcripts, loose
tolerance results, and the full protocol.
This design fixes a low-friction execution regime: literal transcription, a
constructive feasible initialization, and cheap reliable local solves. F is
therefore expected to lead at this pricing point. Certificate-driven tasks,
initialization-poor problems, and regimes where local solves stall increase
the value of FC and can reverse the stage ordering. CPR treats transcription
difficulty, initialization quality, solver reliability, and certificate
value as explicit regime variables.

\paragraph{Results.} All $270$ unique calls shared across nested arms parsed against their schemas with
\ablToolEvents{} tool events. Table~\ref{tab:abl} and Fig.~\ref{fig:abl}
report the utility and cost of each nested stage package.

\emph{Formal transcription with deterministic solver execution is the most
effective package in this regime.}
It lifts usable yield from $\ablDusable\%$ to $\ablFusable\%$
(\ablPairedFwins{} F wins, \ablPairedDwins{} D wins, and
\ablPairedTies{} ties; two-sided exact sign test $p=\ablSignP$ after excluding
ties) and lowers every reported cost per usable solution
(Table~\ref{tab:abl}).
\par\newpage
F exactly transcribes all 90 programs; deterministic
execution totals $\ablSolverTotalF$s and has zero output dispersion.
The remaining unusable cases are
feasible local solutions with large oracle gaps.

\emph{Initialization is a measured stage-design variable.} Warm-start
sensitivity is $\ablFusable\%$ usable yield from the standardized feasible
origin and $\ablFinitRandom\%$ across seeded random starts, attributing
$17.0$ percentage points of usable yield to initialization
(App.~\ref{app:abl-details}).

\emph{Convexification exposes two stage-specific error modes.} Adding the
convexification commitment changes yield to $\ablFCusable\%$.
SDR is usable on all tight cases ($\ablSDRtightU/\ablSDRtightN$) and
infeasible on all measurably loose cases
($\ablSDRlooseFeas/\ablSDRlooseN$), exposing relaxation leakage. OSM is
feasible in $\ablSCAfeasN/\ablSCAN$ cases but usable in
$\ablSCAusableN/\ablSCAN$, exposing conservative over-contraction. Detection
matches the indefinite set in $90/90$ calls; strategy varies on
$\ablStratMixedPids/30$ problems. Per-problem usable counts give $25/0/5$
F wins/FC wins/ties, quantifying the cost of the surrogate commitment in this
regime.

\emph{Verification raises returned-answer precision.} FCV returns
$\ablFCVreturnedN/90=63.3\%$ of answers with $100\%$ feasibility, versus
$\ablFCprecision\%$ for FC, and changes usable yield from $\ablFCusable\%$ to
$\ablFCVusable\%$. The transferred gate covers
$\ablGateCoveredN/\ablGateN$ endpoints ($\ablGateCoverage\%$). F retains the
highest yield at $\ablFusable\%$; the ablation prices the committed
convexification stage through its observed relaxation leakage,
over-contraction, and execution cost. FCV versus FC gives $1/0/29$
wins/losses/ties by problem, localizing the recovery to one problem.

\begin{table}[!t]
\begin{center}
\small
\setlength{\tabcolsep}{3.4pt}
\begin{tabular}{lcccccc}
\toprule
Arm & calls & usable \% {\scriptsize [CI]} & ret./prec.\ \% &
in/use (k) & out/use (k) & wall/use (s) \\
\midrule
Direct (D) & 1 & 41.1\,{\scriptsize [29,52]} & 100.0 / 46.7 & 31.0 & 14.6 & 610 \\
Formalize (F) & 1 & 90.0\,{\scriptsize [77,100]} & 100.0 / 100.0 & 14.4 & 1.1 & 39 \\
$+$Convexify (FC) & 2 & 20.0\,{\scriptsize [9,32]} & 100.0 / 62.2 & 127.4 & 7.8 & 327 \\
$+$Verify/repair (FCV) & 2 & 21.1\,{\scriptsize [10,33]} & 63.3 / 100.0 & 120.7 & 7.4 & 310 \\
\bottomrule
\end{tabular}

\end{center}
\caption{Path ablation ($30 \times 3$ per arm), against ground truth.
Calls $=$ LLM calls/sample; ret./prec.\ gives returned coverage and feasibility
among returned answers; usable $=$ $10^{-6}$-feasible with gap
$\le 0.05R$ to the best-known oracle (SDR-certified global on
$\ablOracleCertN/30$ problems; per-problem bootstrap CIs;
App.~\ref{app:abl-details}). Cost columns report input tokens, output tokens,
and total LLM$+$executor wall-clock per usable output. The $270$ unique calls
comprise $90$ D, $90$ F (shared by F/FC/FCV), and $90$ C (shared by FC/FCV).}
\label{tab:abl}
\end{table}

\section{Experiments III: feedback-conditioned action changes}
\label{sec:exp-ent}

We evaluate one feedback transition on two committed states per problem.
Actions keep/repair the endpoint (A), switch SDR$\leftrightarrow$OSM (B),
roll back to the formal non-convex program (C), or abstain (D). In
\entCalls{} independent pre/post calls, feedback adds execution status,
residual, objective, SDR bound, and gap.

\begin{table}[t]
\centering
\small
\setlength{\tabcolsep}{4pt}
\begin{tabular}{lccc}
\toprule
Metric & state only & state $+$ feedback & paired change [95\% CI] \\
\midrule
Reported allocation entropy (nats) & \entHPre{} & \entHPost{} &
  $\entHDelta{}$ decrease $[\entHCILo{},\entHCIHi{}]$ \\
Reported usable allocation mass & \entUMPre{} & \entUMPost{} &
  $+\entUMDelta$ pp $[\entUMCILo{},\entUMCIHi{}]$ \\
Reported feasible allocation mass & \entFMPre{} & \entFMPost{} &
  $+\entFMDelta$ pp $[\entFMCILo{},\entFMCIHi{}]$ \\
Selected-action usable yield & \entSelUsePre{} & \entSelUsePost{} &
  $+\entSelUseDelta$ pp $[\entSelUseCILo{},\entSelUseCIHi{}]$ \\
\bottomrule
\end{tabular}
\caption{Paired-condition experiment (\entPairs{} states, \entCalls{} independent calls).
Entropy is computed from schema-elicited action allocations; usability comes
from deterministic offline execution against ground truth. Paired changes and
confidence intervals are computed from unrounded values, then rounded for
display.}
\label{tab:entropy}
\end{table}

The model selects one action for ground-truth execution and reports an A--D
allocation for the interface-concentration diagnostic. Primary outcomes are
selected-action change and executed utility; bootstrap intervals resample
problems.

Feedback changes \entChanged{}/\entPairs{} selections:
\entChangedGood{} improve usability, \entChangedNeutral{} are neutral, and
\entChangedBad{} are harmful. Selected usable yield rises
\entSelUsePre{}$\to$\entSelUsePost{}. Always-C and a rule that keeps only a
feasible endpoint within $0.05R$ of the SDR bound both reach $54/60=90.0\%$,
versus $36.7\%$ for the selector. Their estimated executor totals are
$0.042$\,s and $0.034$\,s, while $60$ post-feedback controller calls use
$633.5$\,s, exposing a decision-quality and controller-cost gap.

Reported allocation entropy decreases by \entHDelta{} nats and usable mass
rises by \entUMDelta{} points (Table~\ref{tab:entropy}); nevertheless,
\entUsefulReopen{} entropy-expanding pairs gain usable mass and
\entOverContract{} concentrating pairs lose it. The diagnostic therefore
pairs concentration with executed utility at a fixed action interface.

\section{Discussion and conclusion}
\label{sec:disc}

\paragraph{Self-proposal and executable validation.}
A \egCalls-output pilot elicits provisional states within the receiving call
from the natural-language problem alone (App.~\ref{app:endogenous-pilot}).
Two-repeat collision entropy is \egHDirect{} nats in both conditions and
usable yield is \egDirectUse{} direct versus \egEndoUse{} with self-proposal;
deterministic evaluation confirms \egCheckCorrect{}/\egCheckClaims{} recorded
endpoint checks. This boundary assigns proposals to the model and trust to
executable feedback.

\paragraph{Transplanted-state concentration and utility.} On six problems,
receiving calls condition on another trajectory's surrogate state. Endpoint
entropy falls \epHZero{}$\to$\epHLate{} ($\Delta H=\epHDelta{}$
$[\epHCILo{},\epHCIHi{}]$), while acceptable mass falls
\epMZero{}$\to$\epMLate{} despite feasibility rising
\epFZero{}$\to$\epFLate{} (App.~\ref{app:endpoint-probe}). The transplant
concentrates feasible, objective-poor endpoints, realizing the predicted
provenance-sensitive failure mode of an unverified hard commitment.

\paragraph{Implications and takeaways.} In the measured low-friction regime,
F is the highest-value package, FC exposes relaxation leakage and
over-contraction, and random starts assign 17 yield points to initialization.
For pipelines with committed executable interfaces, compare shared-call
nested packages with stage-local checks, and price a transformation against
the target distribution's leakage and over-contraction rates. Reserve hard
constraints for trusted or executor-verified invariants; keep model-proposed
states soft and reversible until validation. A continuous
residual maps a repair budget to a cutoff; repair and retry prices select the
policy (here $c_r/c_v=0.098$). Certificate value, initialization, solver
reliability, and the cost vector index the audit.

\paragraph{Conclusion and reproducibility.}
\label{sec:conc}
CPR pairs a source-aware path hypothesis with an executable stage audit. The
case study prices formalization, surrogate commitment, verification, and
feedback; the endpoint probes show how provenance and executable validation
govern commitment strength. Fixed-seed scripts and logged outputs reproduce
every analysis; the supplement contains code, prompts, schemas, data, logs,
and collection audits.

\bibliography{references}
\bibliographystyle{colm2026_conference}

\appendix

\paragraph{Public reproducibility release.}
Code, processed experimental records, figure-generation scripts, prompts, and
schemas are available at \url{https://github.com/RedForestLonvor/Constrained-Path-Reasoning}.

\section{Stage-interface measurement map}
\label{app:diagrams}

Table~\ref{tab:mapping} connects each CPR measurement dimension to its case-study
operationalization and empirical evidence.

\begin{table}[!h]
\begin{center}
\scriptsize
\renewcommand{\arraystretch}{0.92}
\begin{tabular}{@{}>{\raggedright\arraybackslash}p{0.23\linewidth}
                    >{\raggedright\arraybackslash}p{0.31\linewidth}
                    >{\raggedright\arraybackslash}p{0.36\linewidth}@{}}
\toprule
Measurement dimension & Case-study operationalization & Empirical evidence \\
\midrule
Transition factorization & NL $\to$ program $\to$ surrogate $\to$ solution
stages & per-stage exactness and stage-package cost/yield
(Tables~\ref{tab:abl}, \ref{tab:abl-stages}) \\
State-space contraction & candidate-level $q_\theta^{\mathcal C_k}$;
fixed A--D allocation as a local instantiation & paired-condition allocation
change and executed action utility (Table~\ref{tab:entropy}); fixed
endpoint entropy and acceptable mass under state transplantation
(App.~\ref{app:endpoint-probe}); collision diagnostic under elicited
self-proposal (App.~\ref{app:endogenous-pilot}) \\
Regularity induction & numerical convex-surrogate solve; SDR lower bounds and
dual diagnostics where available & solver success; SDR lower bounds (oracle certified on
$\ablOracleCertN/30$) \\
Feedback & endpoint residual $\res(x_c)$ & residual--projection-distance fit, gate
coverage, triage yield/precision (Table~\ref{tab:triage}) \\
\bottomrule
\end{tabular}
\end{center}
\caption{Theory-to-measurement map. Fixed action interfaces measure local
concentration; endpoint interfaces probe state transplantation and elicited
self-proposal. Regularity, feedback, localized error, and cost use executor
outputs.}
\label{tab:mapping}
\end{table}

\section{Continuous constrained-path formulation}
\label{app:continuous}

Let $\phi:\R^d\times[0,T]\to\R^m$ be continuously differentiable and
let $\mathcal M_t^{\mathrm{inv}}=\{z:\phi(z,t)=0\}$ encode a prescribed or
independently certified physical, logical, or geometric invariant. A
pre-learned manifold enters this hard set after the same validation. Assume
the Jacobian
$J_t=D_z\phi(z_t,t)$ has full row rank along the path. Given a base drift
$v_t=f_0(z_t,t,x)+g_\theta(z_t,t,x)$, combining a known prior ODE $f_0$ with
a learned residual $g_\theta$, the least-norm normal correction is
\[
u_t=-J_t^\top(J_tJ_t^\top)^{-1}
\bigl(\partial_t\phi(z_t,t)+J_tv_t\bigr),
\qquad \dot z_t=v_t+u_t.
\]
It gives
$\tfrac{d}{dt}\phi(z_t,t)=J_t\dot z_t+\partial_t\phi=0$;
an initially admissible path therefore follows the moving manifold while
retaining the tangential component of $v_t$. Time-varying inequality sets can
be represented by the corresponding projected differential inclusion using
the normal cone of the admissible set.

For an uncertified learned manifold
$\widehat{\mathcal M}_t=\{z:\widehat\phi_\psi(z,t)=0\}$, CPR uses a finite
attraction such as
$-\lambda_t\nabla_z\|\widehat\phi_\psi(z,t)\|^2/2$, or a confidence tube,
inside the base drift. Validation controls $\lambda_t$ and can promote the
proposal to the hard invariant above. Finite attraction preserves departures,
revision, and competing hypotheses during heuristic search.

This formulation supplies two continuous analogues of the discrete CPR
quantities. First, for a fixed reference measure, an endpoint or latent-state
density $\rho_t$ transported by the tangential field obeys a continuity
equation; on a fixed manifold,
$\partial_t\rho_t+\nabla_{\mathcal M}\!\cdot(\rho_t\dot z_t)=0$ and its
differential-entropy rate is the expected intrinsic divergence. Learned
manifold dimension, divergence, reachable volume, and fixed checkpoint probes
therefore provide distinct concentration diagnostics. Second, if perturbations
satisfy $\tfrac{d}{dt}\norm{e_t}\le L(t)\norm{e_t}+\eta(t)$, Gr\"onwall's
inequality gives
\[
\norm{e_T}\le e^{\int_0^T L(r)dr}\norm{e_0}
+\int_0^T e^{\int_s^T L(r)dr}\eta(s)ds,
\]
the continuous counterpart of Eq.~\eqref{eq:error}. Discrete operator paths
are compositions or numerical discretizations of these flows; hybrid CPR
alternates continuous segments with symbolic or tool-execution commitments.
The endpoint experiments use discrete interfaces. The self-proposal pilot
generates its exposed states within the receiving call; an endogenous
continuous-path experiment would sample checkpoints within each flow
realization and apply the same fixed endpoint probe.

\section{Proof and discussion of Proposition~\ref{prop:triage}}
\label{app:proof}

Throughout, $\F = \{x \in \R^n : g_i(x) \le 0,\ h_j(x) = 0\}$ is
non-empty, $g_i, h_j \in C^1$, and $\res$ is \eqref{eq:res}; distances are
Euclidean. A bounding constraint (in our instances the smooth ball
$\norm{x}^2 - R^2 \le 0$) is among the $g_i$ and enters $\res$, so $\F$ and
each tube $U_{\delta_0}$ are compact,
and every $x_c$ has a (possibly non-unique) projection $x^\star \in
\argmin_{x \in \F} \norm{x - x_c}$ with $\norm{x_c - x^\star} =
\dist(x_c,\F)$. The harness evaluates the ball contribution as
$(\norm{x}-R)_+$. Outside the ball,
$\norm{x}^2-R^2=(\norm{x}-R)(\norm{x}+R)$; on each bounded tube the two
forms are locally equivalent, changing the modulus but preserving the
error-bound exponent.

\paragraph{Parts (i)--(iii).} Since $\res(x_c) \le \delta_0$ by assumption,
\eqref{eq:EB} applies at $x_c$; if additionally $\res(x_c) \le
(\varepsilon/\kappa)^{1/\theta}$, then $\dist(x_c,\F) \le \kappa
\res(x_c)^\theta \le \varepsilon$; take $x^\star$ the projection. Part (ii)
is identical with $\beta$ in place of $\varepsilon$. Part (iii) says the
bound becomes non-informative at radius $\beta$:
for $\res(x_c) > (\beta/\kappa)^{1/\theta}$ the bound \eqref{eq:EB} yields
only $\dist \le \kappa\res^\theta$, whose right-hand side exceeds $\beta$.
Part (iii) therefore
marks the boundary of the radius-$\beta$ certificate; \S\ref{sec:exp-cs}
measures realized repair yield on both sides of this boundary. \hfill$\square$

\paragraph{Part (a): the regular regime.} The one-constraint case gives the
core argument. Let a single
inequality $g$ be active with $\grad g \ne 0$ on the boundary; by compactness
and continuity choose $\delta_0>0$ and $\gamma>0$ with $\norm{\grad g(x)} \ge
\gamma$ on the tube $T = \{0 < g \le \delta_0\}$. Fix $x \in T$ with $g(x) =
\rho$ and follow unit-speed steepest descent $\dot u = -\grad g(u)/\norm{\grad
g(u)}$ from $x$: while in $T$, $\tfrac{d}{ds} g(u(s)) = -\norm{\grad g(u(s))}
\le -\gamma$, so $g$ hits zero by arclength $s^\star \le \rho/\gamma$, and
$\dist(x,\F) \le \norm{u(s^\star) - x} \le \rho/\gamma = \res(x)/\gamma$,
i.e.\ \eqref{eq:EB} with $\theta=1$, $\kappa = 1/\gamma$. For general
systems, assume MFCQ holds at every $\bar x\in\F$. At each such point it
implies metric regularity at $(\bar x,0)$ of the \emph{set-valued} constraint
mapping
$M(x)=\big(g(x),h(x)\big)-\big(\R^m_-\times\{0\}\big)$
\citep{robinson1976stability}. Thus there are a neighborhood
$V_{\bar x}$ and modulus $\kappa_{\bar x}$ for which
$\dist(x,\F)\le\kappa_{\bar x}\dist(0,M(x))$ on $V_{\bar x}$.
Finite-dimensional norm equivalence lets the max-residual
\eqref{eq:res} control $\dist(0,M(x))$. Compactness of $\F$ provides a
finite subcover $V_1,\ldots,V_N$; taking the largest of the corresponding
moduli gives a uniform linear residual bound on
$V=\bigcup_{i=1}^N V_i\supseteq\F$
\citep{rockafellar1998variational,pang1997error}. The bound extends from $V$
to the whole tube: on the compact set $U_{\delta_0} \setminus V$ the
continuous residual has a positive minimum $r_{\min} > 0$ (since
$\res^{-1}(0) = \F \subset V$) while $\dist(\cdot,\F)$ is bounded by some
$D < \infty$, so $\dist(x,\F) \le (D/r_{\min})\,\res(x)$ there; taking the
larger of the two moduli gives \eqref{eq:EB} with $\theta = 1$ on all of
$U_{\delta_0}$. Under LICQ the local modulus
can be bounded in terms of the smallest singular value of the active Jacobian;
the exact relation between error-bound moduli and Jacobian spectra depends on
the local nonlinear geometry. \hfill$\square$

\paragraph{Part (b): the quadratic-growth regime.} Assume there are $\mu>0$,
$\delta_0>0$ with $\res(x) \ge \mu \dist(x,\F)^2$ on $U_{\delta_0}$
(quadratic growth of the violation transverse to $\F$; the implication below
is elementary algebra---cf.\ quadratic-growth/error-bound equivalences in
convex settings \citep{drusvyatskiy2018error}). Then $\dist(x,\F) \le \mu^{-1/2}
\res(x)^{1/2}$, i.e.\ \eqref{eq:EB} with $\theta = \tfrac12$, $\kappa =
\mu^{-1/2}$. This exponent is attained: for $g(x) = p(x)^2 \le 0$ with $p \in
C^2$, $\grad p \ne 0$ on $S = \{p = 0\} = \F$, one has $\res = p^2$ while
$\dist(x, S) = |p(x)|/\norm{\grad p} + o(|p(x)|)$ near $S$, so $\dist \asymp
\res^{1/2}$ and no exponent $\theta > \tfrac12$ can hold as $\res \to 0$.
\hfill$\square$

\begin{remark}[Higher-order degeneracy]
\label{rem:not-exhaustive}
Higher-order degeneracy produces other exponents: $g(x) = x^4 \le 0$ on
$[-1,1]$
has $\F = \{0\}$, $\res(x) = x^4$, $\dist(x,\F) = |x| = \res(x)^{1/4}$, so the
sharp \eqref{eq:EB} exponent is $\theta=\tfrac14$; anisotropic degeneracies
can mix exponents by direction. The verifier therefore accepts the
$(\kappa,\theta)$ pair---analytical or calibrated---that describes the target
instance family.
\end{remark}

\begin{remark}[Certificate vs.\ gate]
When $(\kappa,\theta)$ come from verified constraint qualifications, the tier
decisions of Proposition~\ref{prop:triage} are guarantees. When they come from
the empirical calibration of \S\ref{sec:gate}, $(\hat\kappa,\hat\theta)$
define a power-law predictor whose semantics are summarized by held-out
coverage ($95.0\%$ with respect to the projection proxy $\pdist$ here).
\end{remark}

\section{Case-study experimental details}
\label{app:cs-details}

\paragraph{Instance generator.} $c \sim \mathcal N(0, I)$ normalized;
$A_k = Q \operatorname{diag}(\lambda) Q^\top$ with $Q$ Haar-orthogonal and
mixed-sign $\lambda$, negative eigenvalues scaled by a non-convexity factor
$\sim U[1.5, 4.5]$ (at least one negative eigenvalue enforced); $b_k \sim
\mathcal N(0,I)$; $d_k \sim -U[0.5, 3]$ (so $x = 0$ is strictly feasible);
$R \sim U[3,8]$; $n \in \{3,4,5,6,8\}$, $m \in \{3,4\} \cdot n$. SDR solved
with CLARABEL (SCS fallback) via CVXPY \citep{diamond2016cvxpy}.

\paragraph{Projection proxy and repair.} For starts $x_c$, $0$, and $8$
random points, let $\tilde x_\ell\in\F$ be each feasible SLSQP projection
return. We use
$\pdist(x_c,\F)=\min_\ell\norm{x_c-\tilde x_\ell}_2\ge\dist(x_c,\F)$.
Repair uses projected subgradient descent on
$\Phi(x) = \sum_k \max(0, q_k(x))^2 + \max(0, \norm{x} - R)^2$ with fixed step
$2R/160$; the cumulative path length at the first numerically feasible iterate is
recorded per instance, so success at any radius budget $\beta$ is evaluable
post hoc without re-running.

\paragraph{Fitting and calibration.} Log--log least squares of $\pdist$ on
$\res$ over instances with $\res > 10^{-6}$; $95\%$ CIs from $2{,}000$-sample
bootstrap; $\hat\kappa$ is the $95$th percentile of
$\pdist/\res^{\hat\theta}$
on the calibration split. Experiment sizes: E1 $380$ generated / $325$ with
$\res>0$; degenerate family $40$ instances $\times\, 24$ perturbation
magnitudes; verification $360$ held-out instances; scaling $4 \times 110$.
This gives $1{,}180$ generated QCQPs plus $40$ engineered degenerate
polynomial instances, and $2{,}140$ evaluated endpoints (counting the 24
perturbations for each degenerate problem); the full suite
takes $\approx47$\,s on a laptop CPU. Seeds are fixed in
\texttt{run\_experiments.py}.

\paragraph{Projection-density sensitivity.} An independent fixed-seed bank
of $80$ generated QCQPs reuses each SDR endpoint while increasing the total
projection starts from $10$ to $50$ and $100$. Among the $67$ violated
endpoints used in each fit, the slopes are $1.02397$, $1.02375$, and
$1.02375$, with $R^2=0.963$, $0.963$, and $0.963$. The $100$-start protocol
improves $11/80$ projection proxies relative to $10$ starts; the median and
mean ratios $\pdist_{100}/\pdist_{10}$ are $1.000$ and $0.995$. This stable
within-bank result assigns the small superlinear offset primarily to pooled
cross-system scaling and the finite numerical range. The archive includes
the script, all per-instance proxies, and the three fitted summaries.

\paragraph{Calibration support (\S\ref{sec:gate}).} The fit-set residuals
span $[r_{\min}, r_{\max}] = [1.05\times10^{-6},\, 56.7]$, containing every
held-out infeasible endpoint and all $34$ infeasible LLM-formalized endpoints.
The $61$ endpoints below $r_{\min}$ are already numerically feasible
($\res\le10^{-6}$). Thus every accept/repair decision on an infeasible
endpoint uses the power law within calibration support.

\begin{table}[h]
\begin{center}
\begin{tabular}{lcccc}
\toprule
$n$ & $4$ & $6$ & $8$ & $12$ \\
\midrule
$\hat\theta$ & $1.053$ & $1.080$ & $1.053$ & $1.063$ \\
$\hat\kappa_{\mathrm{ls}}$ (least-squares $K$) & $0.152$ & $0.151$ & $0.137$ & $0.127$ \\
$R^2$ & $0.977$ & $0.987$ & $0.992$ & $0.995$ \\
\bottomrule
\end{tabular}
\end{center}
\caption{Scaling of the fitted residual--projection-distance relation with dimension
($m = 3n$, $\ge 75$ violated instances per column). The fitted exponents are
$1.053$--$1.080$ and the constants are $0.127$--$0.152$ across dimensions.}
\label{tab:scaling}
\end{table}

\begin{figure}[h]
\centering
\includegraphics[width=0.86\linewidth]{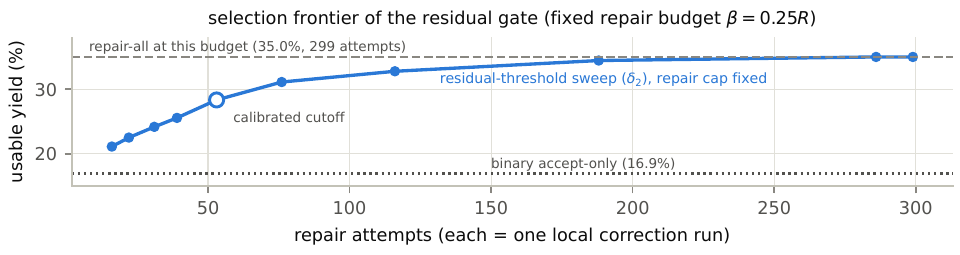}
\caption{Residual-ranking frontier with the operational repair-path cap fixed
at $\beta=0.25R$. The curve sweeps only the selection cutoff $\delta_2$;
every selected endpoint receives the same repair cap. The marked calibrated
cutoff is $\delta_2=(\beta/\hat\kappa)^{1/\hat\theta}$.}
\label{fig:frontier}
\end{figure}

\paragraph{Operating point of Table~\ref{tab:triage}.} $\varepsilon = 0.02R$,
$\beta = 0.25R$; $\delta_1 = (\varepsilon/\hat\kappa)^{1/\hat\theta}$,
$\delta_2 = (\beta/\hat\kappa)^{1/\hat\theta}$. Tier counts at this point:
$77$ \textsc{near-feasible} (of which $61$ already numerically feasible), $37$
\textsc{repair}, $246$ \textsc{reject}. Repair-all and the gated policy use
the identical repair operator and budget; differences are purely which
endpoints they attempt. Sweeping budget and gate \emph{together}
produces a separate budget-sensitivity curve. Figure~\ref{fig:frontier}
instead fixes the operational cap at $\beta=0.25R$ and varies only the
selection threshold $\delta_2$; its marked point uses
$\delta_2=(\beta/\hat\kappa)^{1/\hat\theta}$.

\paragraph{Break-even cost model.} Let $c_r$ be the price of a repair
attempt and $c_v$ the expected downstream price of retrying or
re-convexifying an unresolved endpoint. Repair-all makes $299$ attempts and
leaves $234$ endpoints unresolved; the gate makes $53$ attempts and leaves
$258$ unresolved. Hence
\[
C_{\rm all}=299c_r+234c_v,\qquad
C_{\rm gate}=53c_r+258c_v.
\]
The gate has lower total price precisely when
$246c_r>24c_v$, or $c_r/c_v>0.09756$. This boundary turns the observed
cost--yield frontier into a deployment rule: repair-all covers the
low-repair-price region, and residual triage covers the
high-repair-price/high-retry-value region.

\section{Path-ablation protocol and additional results}
\label{app:abl-details}

\paragraph{Problem bank.} $30$ problems ($n \in \{2,3,4\}$, $10$ each,
$m = 3n$), generated from the \S\ref{sec:exp-cs} family with all coefficients
rounded to three decimals; the rounded values are the ground truth and appear
verbatim in the NL text (Fig.~\ref{fig:nl-example}). By construction $x = 0$
is strictly feasible. Reference values per problem: best-known objective from
a $26$-start SLSQP oracle on the true problem (unchanged under $100$ starts
on all $30$ problems; the SDR lower bound certifies global optimality on
$\ablOracleCertN/30$), and the true-SDR residual as a looseness marker
($19/30$ problems have loose SDR). On the $\ablOracleCertN$ SDR-certified
problems, D/F/FC/FCV usable yield is $52.8/100.0/50.0/52.8\%$; on the $18$
problems evaluated against best-known multi-start values it is
$33.3/83.3/0.0/0.0\%$. The FC/FCV failures concentrate on loose-SDR
problems. Precisely, $18/19$ loose-SDR problems belong to the uncertified
subset, all $18/18$ uncertified problems have loose SDR, and the remaining
loose-SDR problem (ID 18) is SDR-certified.

\begin{figure}[h]
\small
\begin{center}
\fbox{\begin{minipage}{0.93\linewidth}\ttfamily\footnotesize
A planner chooses 2 decision variables x1..x2.\\
They want to MINIMIZE the cost f(x) = 0.186*x1 - 0.983*x2.\\
The choice must satisfy ALL of the following constraints:\\
\;\;(C1)\; -0.967*x1\^{}2 - 6.875*x1*x2 - 3.995*x2\^{}2 + 0.846*x1 + 0.063*x2 - 1.424 <= 0\\
\;\;\dots\ (five more quadratic constraints)\ \dots\\
\;\;(C7)\; the Euclidean norm bound sqrt(x1\^{}2 + ... + x2\^{}2) <= 6.0
\end{minipage}}
\end{center}
\caption{Example natural-language problem (abridged; problem 0 of the bank).}
\label{fig:nl-example}
\end{figure}

\paragraph{LLM configuration.} All calls:
\texttt{codex exec --sandbox read-only --ephemeral -m gpt-5.5 -c
model\_reasoning\_effort="medium" --json --output-schema <schema>}. The prompt
prohibits tool use and demands a single JSON object; the JSONL event logs
confirm zero command/tool events across all calls (CLI
\texttt{codex-cli 0.144.1}; calls issued 2026-07-11 to 07-13). The CLI uses
provider-default stochastic sampling; the three samples per problem are
repeated calls, and statistical comparisons use problems as units. Each
problem $\times$ sample makes one call per LLM stage: Direct
(arm D), Formalize (shared by arms F/FC/FCV), Convexify-given-formalization
(shared by FC/FCV); $270$ calls total. Raw event logs, final messages, parsed
outputs, and per-call token/wall-clock accounting are all in the supplementary
material.

\paragraph{Deterministic executors.} Arm F: SLSQP from the origin on the
formalized non-convex program ($300$ iterations); the origin is a warm start
($x=0$ feasible by construction), and from seeded random starts in the ball
the same executor's usable rate is $\ablFinitRandom\%$ (mean over $10$
starts per problem) vs.\ $\ablFusable\%$ from the origin
(\texttt{local\_robustness.py}). Arm FC: strategy applied to
the model's formalization---SDR via the \S\ref{sec:exp-cs} lifting;
``SCA'' as one SCA-style inner majorization (OSM) anchored at $x_0=0$ by
dropping each quadratic's concave part; results target the single committed
surrogate. ``Already convex'' is valid when every $P_k$ is positive
semidefinite, with other selections recorded as strategy errors. Arm FCV:
triage with the
\S\ref{sec:exp-cs} gate ($\hat\kappa = 0.350$, $\hat\theta = 1.046$,
$\varepsilon = 0.02R$, $\beta = 0.25R$) on the model's own formalization;
gated infeasible endpoints get bounded repair, and the post-repair residual
check returns only feasible endpoints.

\paragraph{Evaluation.} Ground-truth residual \eqref{eq:res}, feasibility
at tight ($10^{-6}$) and loose ($10^{-3}$) tolerances, and the signed
objective gap $f(x) - f_{\mathrm{bk}}$ to the best-known oracle value;
usable $=$ numerically feasible at $10^{-6}$ and gap $\le 0.05R$. The
$\ablBeatOracleN/270$ sub-oracle gaps are all ${<}6\times10^{-7}$ and are
treated as numerical ties. Table~\ref{tab:abl-stages} reports
usable rates at gap thresholds $0.02R/0.05R/0.1R$; the arm ordering is
unchanged.
Formalization exactness:
canonicalized coefficient match within $10^{-6}$. Detection exactness:
claimed non-convex index set equals the indefinite set of the model's own
formalization. Dispersion: mean pairwise $\ell_2$ distance between the final
answers of the $3$ samples, divided by $R$.

\begin{figure}[h]
\centering
\includegraphics[width=\linewidth]{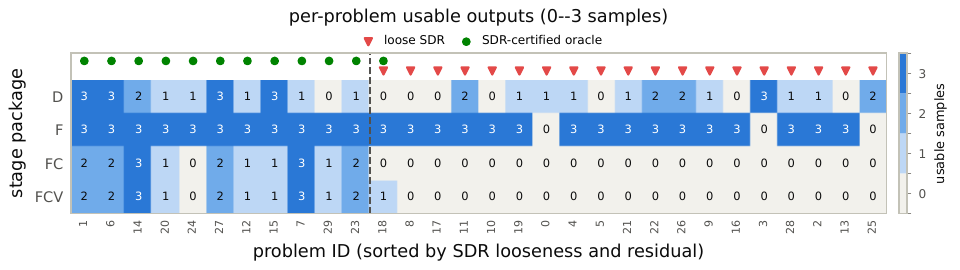}
\caption{Per-problem stage-package outcomes. Each cell is the number of usable
outputs among three samples. Problems are ordered with tight SDRs first, then
loose SDRs, and by true-SDR residual within each group; the dashed line marks
the group boundary. Markers identify loose SDRs and problems whose oracle
value is certified by the SDR lower bound.}
\label{fig:abl-heatmap}
\end{figure}

\paragraph{Per-stage measurements.} Table~\ref{tab:abl-stages} isolates each
LLM stage against its own exactly-checkable target (formalization against the
embedded ground-truth coefficients; detection against the indefinite set of
the model's own formalization), separating model error from error injected by
the committed transitions themselves. Figure~\ref{fig:abl-heatmap} localizes
the resulting package utility by problem and SDR regime.

\begin{table}[h]
\begin{center}
\small
\begin{tabular}{lc}
\toprule
Stage measurement & value \\
\midrule
Direct: parse rate & 100.0\% \\
Formalize: parse / valid / exact-match rate & 100.0\% / 100.0\% / 100.0\% \\
Formalize: harness execution rate & 100.0\% \\
Convexify: parse / detection-exact rate & 100.0\% / 100.0\% \\
Convexify: harness execution rate & 100.0\% \\
Strategy choices (SDR / OSM / already-convex) & 52 / 38 / 0 \\
FCV tiers (feasible / near-feas.\ / repair / reject / no cand.) & 56 / 1 / 2 / 31 / 0 \\
FCV repairs attempted / successful & 3 / 1 \\
FC usable/n by cell: SDR$\times$tight, SDR$\times$loose, OSM & 18/18, 0/34 (feas.\ 0), 0/38 (feas.\ 38) \\
Median gap of feasible answers (D/F/FC/FCV) & 0.01 / $<0.01$ / 1.48 / 1.40 \\
Wall-clock (LLM$+$solver) per usable, s (D/F/FC/FCV) & 610 / 39 / 327 / 310 \\
Input tokens per usable, k (D/F/FC/FCV) & 31.0 / 14.4 / 127.4 / 120.7 \\
Usable \% at gap $\le 0.02R/0.05R/0.1R$: D & 34.4 / 41.1 / 42.2 \\
\quad F & 80.0 / 90.0 / 90.0 \\
\quad FC & 20.0 / 20.0 / 23.3 \\
\quad FCV & 21.1 / 21.1 / 24.4 \\
Matched dispersion, FCV-returned pairs (FC / FCV, $n$=17) & 0.28 / 0.28 \\
Calibrated-gate coverage on FC endpoints ($n$=34) & 88.2\% \\
Tool-use events across all 270 calls & 0 \\
\bottomrule
\end{tabular}

\end{center}
\caption{Stage-level measurements over all $30 \times 3$ samples. Parse
$=$ output matched the JSON schema; valid $=$ formalization is a well-formed
program of the right shape; exact $=$ coefficient match with ground truth
within $10^{-6}$ after canonicalization.}
\label{tab:abl-stages}
\end{table}

\section{Feedback-condition protocol}
\label{app:entropy-details}

\paragraph{Frozen items and prompts.} We use sample indices $0$ and $1$ from
each of the $30$ path-ablation problems, yielding \entPairs{} formalization
and convexification commitments. Each commitment has two independent
state-only/state-plus-feedback calls. Both prompts share the problem,
committed program, detected indices, strategy, rationale, and action menu.
The post condition adds execution status, endpoint, residual, objective, the
SDR lower bound, and their gap. Oracle values and offline action labels are
reserved for evaluation.

\paragraph{Actions and scoring.} A keeps or repairs the current endpoint; B
executes the other surrogate with the same verifier; C runs SLSQP on the
committed program from its standard feasible origin; D abstains. The harness
executes each action with repair radius $0.25R$. Usable means ground-truth feasibility at
$10^{-6}$ and objective gap at most $0.05R$, as in
\S\ref{sec:exp-abl}. Paired intervals resample the $30$ problem IDs with
replacement (2,000 draws, seed 20260715), retaining both samples and
conditions within a problem.

\paragraph{Calls and parsing.} Calls use the same Codex command in
App.~\ref{app:abl-details}, with an output schema requiring one selected label
and four reported allocations in $[0,1]$. All \entCalls{} medium-effort calls parsed;
every reported allocation sum was $1$, the selected label was always an
argmax, and the JSONL logs contain no command/tool item. The archive also
contains six configuration checks; every reported result uses the $120$
explicit medium-effort calls. Full events, outputs, action outcomes, and
analysis tables are included in the supplement.

\paragraph{Offline policy baselines.} Always-A/B/C/D and an offline oracle
are evaluated from the same deterministic action outcomes. The diagnostic
rule selects A only when the current endpoint is $10^{-6}$-feasible and its
objective is within $0.05R$ of the available SDR lower bound; otherwise it
selects C. It chooses A on 11 states and C on 49, attaining $54/60$ usable and
$60/60$ feasible, equal in utility to always-C and the offline oracle. The
measured F executor costs $0.0630$\,s for $90$ runs, giving estimated totals
of $0.0420$\,s for $60$ always-C actions and $0.0343$\,s for the rule's $49$
C actions. The $60$ post-feedback controller calls take $633.5$\,s before
action execution. These measurements price the interface as well as its
action accuracy.

\section{Cross-trajectory surrogate-state transplantation stress test}
\label{app:endpoint-probe}

This repeated-call probe uses problem IDs $\{0,5,10,15,20,25\}$ from the
released bank, giving two problems at each $n\in\{2,3,4\}$. State $S_0$
contains the natural-language problem. State $S_{K-1}$ contains the same
problem, the sample-0 exact formal program, the committed surrogate endpoint,
and its execution record: status, residual, objective, available SDR bound,
and gap. A shared prompt requests one final numeric endpoint under the same
JSON schema and model configuration. The formal program and surrogate state
were produced by sample 0 of another trajectory and then held fixed across
the independent receiving calls; the prompt permits retaining or revising the
supplied endpoint. The estimand is therefore cross-trajectory state
transplantation. An endogenous-path test instead generates a fresh
provisional state inside every sampled trajectory before measuring its
continuation.

The initial manifest targeted twelve repetitions. A collection-budget
amendment, made after 63 requests and before offline scoring, capped the run
at 80 requests and retained repetitions 0--5 for every problem--state cell.
The resulting Cartesian design has $6\times2\times6=72$ completed calls.
Six in-flight requests interrupted at the amendment were repeated, producing
\epActualRequests{} actual requests in total. The archive retains both
versioned manifests (current SHA-256 prefix \texttt{4432fa5b4522}) and every
attempt record.

For problem $i$, the fixed endpoint encoding divides each coordinate of
$x/R$ into bins of width $0.05$ and includes separate \textsc{invalid} and
\textsc{abstain} symbols. Let $\widehat Q_{ik}$ be the resulting empirical
distribution at state $S_k$. We report plug-in entropy with the
Miller--Madow correction, the number of occupied endpoint bins, and acceptable
mass: the fraction of calls whose endpoint is $10^{-6}$-feasible and within
$0.05R$ of the best-known objective. The primary paired estimands are
\[
\Delta H_i=H(\widehat Q_{i0})-H(\widehat Q_{i,K-1}),\qquad
\Delta M_i=M_{i,K-1}-M_{i0}.
\]
Problem-level paired bootstrap intervals preserve all six calls within each
state. The pre-specified directional criterion is a positive mean
$\Delta H$ together with mean $\Delta M\ge-0.05$; raw collision counts and
per-problem estimates accompany the aggregate because six samples make
entropy resolution explicit. This stress test reuses the endpoint
representation and utility-preservation criterion of
Hypothesis~\ref{hyp:ambiguity} while intervening on state provenance.

\begin{figure}[h]
\centering
\includegraphics[width=\linewidth]{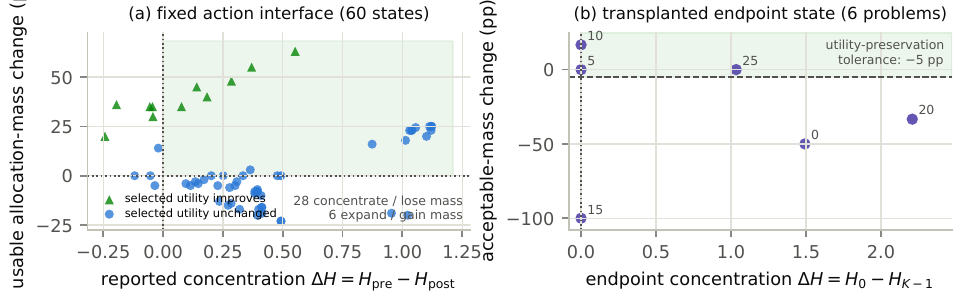}
\caption{Concentration paired with task utility. \textbf{(a)} For the fixed
action interface, many reported allocation distributions concentrate while
usable allocation mass decreases; marker shape records the executed utility
of the selected action. \textbf{(b)} The transplanted endpoint state
concentrates several problem-level endpoint distributions while acceptable
mass falls beyond the pre-specified five-point tolerance. Shading marks the
joint concentration-and-utility region for each interface.}
\label{fig:concentration-utility}
\end{figure}

All \epEffectiveCalls{} completed outputs parse under the schema and contain
zero tool events. Mean corrected entropy changes from \epHZero{} at $S_0$ to
\epHLate{} at $S_{K-1}$, giving $\Delta H=\epHDelta{}$
$[\epHCILo{},\epHCIHi{}]$. The per-problem entropy comparison is
$3/0/3$ wins/losses/ties (one-sided sign test excluding ties, $p=0.125$).
Acceptable mass changes from \epMZero{} to \epMLate{}, or
\epMDeltaPP{} percentage points $[\epMCILoPP{},\epMCIHiPP{}]$, while feasible
mass rises from \epFZero{} to \epFLate{}. The directional criterion is
therefore unmet for the transplanted state: it concentrates the endpoint
distribution and raises feasibility while placing much of its mass outside
the objective-utility tolerance. The receiving calls inherit a surrogate
basin they did not select, localizing the measured failure to hard
cross-trajectory conditioning on an unverified proposal.
Figure~\ref{fig:concentration-utility} places this endpoint result beside the
fixed-action diagnostic on their respective concentration--utility planes.

\begin{table}[h]
\centering
\small
\setlength{\tabcolsep}{4pt}
\begin{tabular}{rrrrrrrr}
\toprule
ID & $n$ & $H_0$ & $H_{K-1}$ & $\Delta H$ & $M_0$ & $M_{K-1}$ & $\Delta M$ \\
\midrule
0  & 2 & 1.492 & 0.000 & 1.492 & .500 & .000 & $-.500$ \\
5  & 4 & 1.580 & 1.580 & 0.000 & .000 & .000 & $.000$ \\
10 & 3 & 1.492 & 1.492 & 0.000 & .000 & .167 & $.167$ \\
15 & 2 & 0.000 & 0.000 & 0.000 & 1.000 & .000 & $-1.000$ \\
20 & 4 & 2.208 & 0.000 & 2.208 & .333 & .000 & $-.333$ \\
25 & 3 & 1.034 & 0.000 & 1.034 & .000 & .000 & $.000$ \\
\midrule
Mean & & 1.301 & 0.512 & 0.789 & .306 & .028 & $-.278$ \\
\bottomrule
\end{tabular}
\caption{Cross-trajectory state-transplant stress test. Entropies use the
Miller--Madow correction; $M$ is acceptable endpoint mass. Each cell contains
six receiving calls.}
\label{tab:endpoint-probe}
\end{table}

\section{Same-call endogenous-state pilot}
\label{app:endogenous-pilot}

We use the same six problems and endpoint encoding as
App.~\ref{app:endpoint-probe}. Both conditions receive only the
natural-language problem. \textsc{Direct} requests an endpoint under the
usual reasoning process. \textsc{Self-proposal} asks the model, within the
same uninterrupted call, to autonomously expose up to three useful states
and then return an endpoint. Every exposed intermediate artifact therefore
originates inside the receiving call. States directly entailed by the problem
may be marked \textsc{verified-hard}; active-set guesses,
numerical anchors, transformations, and other model proposals remain
\textsc{soft}, \textsc{revised}, or \textsc{rolled-back}. The exposed
\texttt{path\_record} makes proposal provenance observable. Its estimand is
behavior under this elicited interface; latent unprompted checkpoints require
model-internal instrumentation.

\paragraph{Clean amendment.} A collection interruption left two drivers
writing the same 48-tag pool, so that pool is released only as an execution
audit. After identifying the write collision, the clean amendment fixed
repetitions 0--1 for all
$6\times2$ problem--condition cells. It retains five nonoverlapping original
responses and seven recovery-driver responses identified by exact logged
wall times and hashes, then collects the twelve missing cells once in an
exclusive-write directory. This outcome-independent rule yields
\egCalls{} valid outputs (\egNewCalls{} new requests), zero tool events, and
manifest SHA-256 prefix \texttt{b1d099e46479}.

\begin{table}[h]
\centering
\small
\begin{tabular}{lcc}
\toprule
Metric & Direct & Same-call self-proposal \\
\midrule
Miller--Madow endpoint entropy & \egHDirect{} & \egHEndo{} \\
Usable yield ($10^{-6}$) & \egDirectUse{} & \egEndoUse{} \\
Usable yield ($10^{-3}$) & \egDirectLoose{} & \egEndoLoose{} \\
Mean wall-clock/call (s) & \egWallDirect{} & \egWallEndo{} \\
Calls exposing a state & $0/12$ & $\egStateCalls/12$ \\
\bottomrule
\end{tabular}
\caption{Clean same-call self-proposal pilot. With two repeats per cell, the
entropy estimate is a collision diagnostic taking only $0$ or $0.943$ nats
per problem--condition cell.}
\label{tab:endogenous-pilot}
\end{table}

The collision comparison gives \egHWins{}/\egHLosses{}/\egHTies{}
concentration wins/losses/ties and equal mean entropy; it therefore leaves
endpoint concentration unresolved at this sample size. The self-proposal
condition records \egSoftStates{} soft states, \egHardStates{} hard
restatements, and \egRevisedStates{} revision. Usable yield is
\egDirectUse{} versus \egEndoUse{}; the loose-tolerance comparison is
\egDirectLoose{} versus \egEndoLoose{}. An independent implementation
recomputes all 21 returned endpoints with zero residual or objective
difference from the main scorer. Among \egCheckClaims{} path records that
explicitly report checking every original constraint at the returned
endpoint, \egCheckCorrect{} passes deterministic evaluation (also at
$10^{-3}$). The observed interface therefore separates endogenous proposal
from validation: dense-QCQP hypotheses become operational constraints after
executor feedback establishes their validity.

\end{document}